\documentclass[10pt,twocolumn,letterpaper]{article}

\usepackage[pagenumbers]{cvpr} 
\usepackage{graphicx}
\usepackage{cuted}
\usepackage{booktabs}
\usepackage{multirow}
\usepackage{bbding}
\usepackage{orcidlink}
\definecolor{cvprblue}{rgb}{0.21,0.49,0.74}

\title{\texorpdfstring{\textcolor{blue}{$L^3$}}{L3}: Scene-agnostic Visua\textcolor{blue}{l} \textcolor{blue}{L}ocalization in the Wi\textcolor{blue}{l}d}

\author{
    Yu Zhang \orcidlink{0009-0005-2353-988X} \quad
    Muhua Zhu \orcidlink{0009-0005-9049-2321} \quad
    Yifei Xue \orcidlink{0000-0002-4443-4367} \quad
    Tie Ji \orcidlink{0009-0005-0822-0600} \quad
    Yizhen Lao\thanks{Corresponding author} 
 \orcidlink{0000-0002-6284-1724} \\
    Hunan University, Changsha, Hunan 410082, China \\
    {\tt\small \{yuzhang, casmyzhu, iflyhsueh, yizhenlao\}@hnu.edu.cn, jietie\_hnu@163.com}
}

\begin{document}

\maketitle
\begin{strip}
    \centering
    \includegraphics[width=\textwidth]{head.pdf}
    \captionof{figure}{\textbf{Comparison between scene-specific and scene-agnostic visual localization paradigms.} \textbf{\textit{\textcolor{red}{Scene-specific}}} methods require extensive offline preprocessing such as reconstruction or per-scene network training. In contrast, our proposed \textbf{\textit{\textcolor{blue}{scene-agnostic}}} visual localization framework \textcolor{blue}{$L^3$} directly estimates query poses via feed-forward coarse localization and PnP refinement, generalizing to novel scenes without requiring scene representations or preprocessing.}
    \label{fig:head_compare}
\end{strip}

\begin{abstract}
  Standard visual localization methods typically require offline pre-processing of scenes to obtain 3D structural information for better performance. This inevitably introduces additional computational and time costs, as well as the overhead of storing scene representations. Can we visually localize in a wild scene \textbf{without} any off-line preprocessing step? In this paper, we leverage the online inference capabilities of feed-forward 3D reconstruction networks to propose a novel map-free visual localization framework \textcolor{blue}{$L^3$}. Specifically, by performing direct online 3D reconstruction on RGB images, followed by two-stage metric scale recovery and pose refinement based on 2D-3D correspondences, \textcolor{blue}{$L^3$} achieves high accuracy without the need to pre-build or store any offline scene representations. Extensive experiments demonstrate \textcolor{blue}{$L^3$} not only that the performance is comparable to state-of-the-art solutions on various benchmarks, but also that it exhibits significantly superior robustness in sparse scenes (fewer reference images per scene).
\end{abstract}    
\begin{figure*}[t!]
    \centering
    \includegraphics[width=\linewidth]{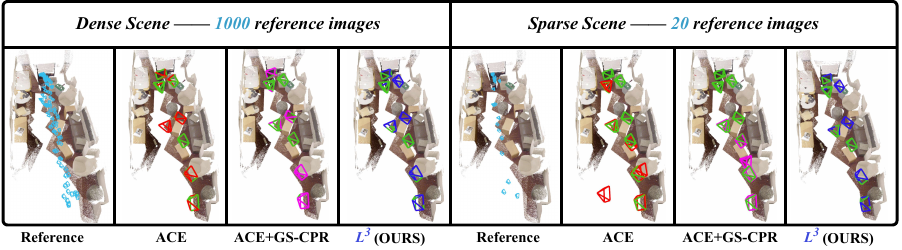}
    \caption{\textbf{Comparison of  dense and sparse scenes.} Green denotes ground truth query poses and references are cyan. The predicted poses by different methods are shown in different colors: red for ACE~\cite{brachmann2023accelerated}, magenta for ACE+GS-CPR~\cite{liugs}, and blue for our proposed \textcolor{blue}{$L^3$}. \textbf{Dense Scene:} Localization using all 1000 reference images. \textbf{Sparse Scene: }We sampled the 1000 reference images and retain only 20. Our \textcolor{blue}{$L^{3}$} significantly outperforms other baselines in this challenging case.}
    \label{fig:introduction_compare}
\end{figure*}

\vspace{-14mm}
\section{Introduction}
\label{sec:intro}
Visual localization, a fundamental task in robotics, autonomous driving, and virtual reality, aims to estimate the 6-DoF pose of a query image from a set of database images with known poses. 

Existing visual localization solutions are mainly categorized into structure-based approaches and image-based approaches:
\textbf{\textit{1)}} Structure-based methods typically store a 3D geometric representation of the scene (3D maps), such as point clouds~\cite{SarlinCSD19,sattler2016efficient,svarm2016city}, meshes~\cite{panek2022meshloc}, NeRF~\cite{chen2024neural,yen2021inerf,zhao2024pnerfloc,zhou2024nerfect,moreau2023crossfire}, 3DGS~\cite{liugs}, or the weights of scene coordinate regression networks~\cite{brachmann2023accelerated,wang2024glace,shotton2013scene,jiang2025r,brachmann2021visual}. By establishing 2D-3D correspondences between the query image and the 3D map, the camera pose can be estimated using the Perspective-n-Point (PnP) solver. 
\textbf{\textit{2)}} In contrast, image-based methods represent the scene as a database of images with poses instead of 3D maps. These approaches rely on image retrieval for the pose approximation~\cite{arandjelovic2016netvlad,berton2023eigenplaces,keetha2023anyloc,berton2025megaloc} or on directly regressing the absolute pose through neural architectures~\cite{kendall2015posenet,shavit2021learning,chen2022dfnet,chen2024map,lin2024learning}. In addition,~\cite{jiang2026imloc} proposed to enhance image databases with local depth maps, enabling the establishment of 2D-3D correspondences for PnP without explicitly reconstructing a global 3D structure.

However, structure-based methods generally achieve superior accuracy but require reconstructing and storing a 3D map of the scene before querying. 
Image-based methods circumvent explicit 3D reconstruction but also require network training (APR) or depth map prediction (depth-map augmentation) before the query phase.
\textbf{In summary}, existing visual localization approaches generally mandate \textbf{scene-specific} offline preprocessings, which are usually time-consuming and computationally expensive. Given this, we raise a core question: \textbf{\textit{Does robust visual localization necessarily require scene-specific priors, such as pre-built maps or optimized networks?}}

To address these challenges, we propose \textcolor{blue}{$L^3$}, a paradigm-shifting framework for scene-agnostic visual localization, as shown in \cref{fig:head_compare}. 
Building upon the inference capabilities of the feed-forward 3D reconstruction network $ \pi^3$~\cite{wang2025pi}, our framework jointly processes a query image and a set of retrieved reference images to perform direct online reconstruction.
By leveraging the strong generalization power of the model, \textcolor{blue}{$L^3$} generates high-quality camera poses and dense geometry in a single pass, completely \textbf{bypassing any scene-specific offline training or pre-built 3D maps}. 
However, the resulting initial poses and point clouds inherently lack a consistent metric scale. To resolve this, we propose a two-stage scale recovery strategy that combines local geometric consistency with global trajectory constraints. Following scale restoration, we perform structure-only bundle adjustment (BA) to optimize the 3D point clouds, fixing the ground-truth poses of the reference cameras. Finally, the refined 6-DoF pose of the query image is recovered through guided matching and a PnP solver between the query image and the optimized 3D structure.

In contrast to existing methods that rely on extensive pre-processing, \textcolor{blue}{$L^{3}$} achieves high localization accuracy while entirely eliminating the necessity for scene-specific offline training or pre-built 3D maps. Furthermore, as demonstrated in \cref{fig:introduction_compare}, our approach not only delivers competitive performance in dense scenarios but also exhibits remarkable robustness under challenging sparse conditions. Consequently, the only required offline processing is the retrieval of reference images, facilitating \textbf{instant deployment in uncharted environments}. Our codes will public online.
Our contributions are summarized as follows:

\begin{itemize}
\item We propose \textcolor{blue}{$L^{3}$}, a scene-agnostic framework for visual localization. To the best of our knowledge, it is the first to achieve performance comparable to state-of-the-art (SOTA) methods without any offline scene-specific optimization or pre-3D-mapping.

\item We design a novel coarse-to-fine localization pipeline featuring a two-stage scale recovery strategy and a structure-only BA which enables accurate metric scale restoration.

\item Extensive evaluations on multiple indoor and outdoor benchmarks demonstrate that \textcolor{blue}{$L^{3}$} generalizes effectively across diverse environments. It not only excels in standard dense-view settings but also exhibits superior robustness in sparse-view scenarios, significantly outperforming existing methods under extreme data constraints.
\end{itemize}
\section{Related Work}
\noindent\textbf{Structure-based methods.} Conventional structure-based methods~\cite{SarlinCSD19,sattler2016efficient,sarlin2020superglue} rely on feature matching and 3D point clouds reconstructed via Structure from Motion (SfM) to derive 2D-3D correspondences for pose estimation, yet they face unavoidable storage overhead despite techniques like point cloud sparsification~\cite{camposeco2019hybrid,yang2022scenesqueezer} or descriptor compression~\cite{dong2023learning,laskar2024differentiable}. Scene Coordinate Regression (SCR) avoids explicit maps by training scene-specific networks to directly regress 3D coordinates, evolving from early random forests~\cite{shotton2013scene,cavallari2017fly,cavallari2019real} to differentiable architectures like DSAC*~\cite{brachmann2017dsac,brachmann2021visual} and efficient decoupled models like ACE~\cite{brachmann2023accelerated}. Recent SCR extensions such as GLACE~\cite{wang2024glace} and R-SCoRe~\cite{jiang2025r} further enhance robustness in large-scale environments through feature diffusion or covisibility constraints. Alternatively, novel view synthesis (NVS) follows a render-and-compare paradigm, where NeRF-based approaches optimize photometric loss~\cite{yen2021inerf,Lin_2021_ICCV}, integrate radiance fields into Monte Carlo filters~\cite{10160782}, or power data augmentation~\cite{moreau2022lens} and pose refinement~\cite{chen2024neural}. Latest works leverage internal radiance representations for feature-level matching~\cite{zhou2024nerfect,zhao2024pnerfloc}.Concurrently, 3D Gaussian Splatting (3DGS) has been utilized for one-shot pose estimation~\cite{matteo20246dgs}. Furthermore, frameworks like GS-CPR~\cite{liugs} leverage 3DGS alongside foundation models such as MASt3R~\cite{leroy2024grounding} to establish robust correspondences.

\noindent\textbf{Image-based methods.} Image-based methods handle dynamic scene changes without explicit 3D structures~\cite{arandjelovic2016netvlad,berton2023eigenplaces,keetha2023anyloc,berton2025megaloc}, typically utilizing image retrieval to approximate poses for subsequent initialization. Absolute Pose Regression (APR)~\cite{kendall2015posenet,shavit2021learning,chen2022dfnet,chen2024map,direct-posenet} maps images directly to 6-DoF poses in a single pass, yet remains limited by lower accuracy and the requirement for scene-specific training~\cite{sattler2019understanding,liu2024hr}. To bridge accuracy and flexibility, ImLoc~\cite{jiang2026imloc} constructs 2D-3D correspondences via depth-map augmentation instead of global models, though it still necessitates triangulation-based depth estimation for all reference images prior to querying.

\noindent\textbf{Feed-forward 3D Reconstruction.} Feed-forward models like Dust3R~\cite{wang2024dust3r} and MASt3R~\cite{leroy2024grounding} enable direct camera pose estimation and scene reconstruction from image sequences, though they often encounter high global alignment costs and stability issues at scale. To mitigate these constraints, Fast3R~\cite{Yang_2025_CVPR} introduces multi-view parallel processing to handle large-scale inputs, while VGGT~\cite{wang2025vggt} and $\pi^3$~\cite{wang2025pi} leverage multi-task pre-training on massive datasets to achieve SOTA accuracy and robustness. In particular, $\pi^3$ provides a permutation-invariant framework with affine-invariant pose predictions.
\section{Methodology}
\noindent\textbf{Problem Statement.} Given a query image $I_q \in \mathbb{R}^{H \times W \times 3}$ with camera intrinsics $\mathbf{K}_q \in \mathbb{R}^{3 \times 3}$, we assume that the top-$K$ candidate reference images $\{I_{r,i}\}_{i=1}^K$ have been retrieved. Our goal is to estimate the absolute 6-DoF query pose  $\mathbf{P}_q = [\mathbf{R}_q | \mathbf{t}_q] \in SE(3)$ by leveraging the reference images $\{I_{r,i}\}_{i=1}^K$, their corresponding poses $\mathbf{P}_{r,i} = [\mathbf{R}_{r,i} | \mathbf{t}_{r,i}]$ and their intrinsics $\mathbf{K}_{r,i}$. We use $(\cdot)^{GT}$ to denote ground truth, and $(\cdot)^{local}$ for network predictions in the local coordinate system. The subscripts $(\cdot)_r$ and $(\cdot)_q$ indicate reference and query images, respectively.

\begin{figure*}[t!]
  \centering
  \includegraphics[width=\textwidth]{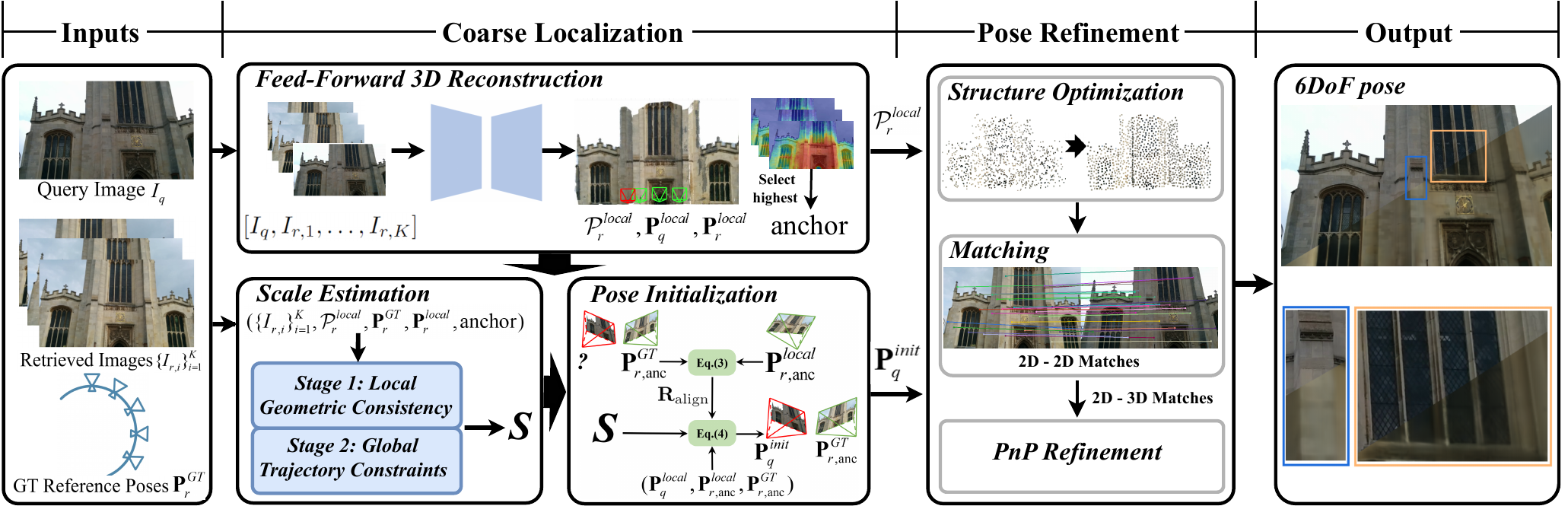} 
  \caption{\textbf{Overview of \textcolor{blue}{$L^3$}.} \textbf{Coarse Localization:} Given a query image $I_q$ and retrieved references $\{I_{r,i}\}_{i=1}^K$, we first perform feed-forward 3D reconstruction to predict local point clouds $\mathcal{P}_{r}^{local}$, query pose $\mathbf{P}_{q}^{local}$ and reference poses $\mathbf{P}_{r}^{local}$. A scale estimation module then computes the scale factor $S$ via a two-stage strategy to initialize the pose $\mathbf{P}_q^{\text{init}}$. \textbf{Pose Refinement:} The pose is refined via structure optimization and PnP to yield the final 6-DoF pose $\mathbf{P}_q$. \textbf{Output:} We compare the query image (top-left) with its rendering under the predicted pose (bottom-right), separated by a diagonal line.}
  \label{fig:pipeline}
\end{figure*}

\noindent\textbf{Framework.} As illustrated in \cref{fig:pipeline}, we introduce \textcolor{blue}{$L^{3}$}, a scene-agnostic framework that estimates absolute camera poses using only retrieved reference images and their associated camera parameters. The process begins in \cref{sec:coarse_localization} with a coarse localization module that jointly processes the query and reference images via a 3D feed-forward reconstruction network to produce initial poses and dense geometric predictions. To resolve the scale ambiguity inherent in these feed-forward outputs, we implement a robust two-stage metric scale estimation strategy and utilize the estimated scale to compute the initial global pose of the query image. Finally, in \cref{sec:pose_refinement}, the scale-corrected 3D structure is optimized through structure-only refinement, followed by a PnP solver to determine the final 6-DoF query pose.

\subsection{Coarse Localization}
\label{sec:coarse_localization}
\noindent\textbf{Feed-Forward 3D Reconstruction.} In this stage, we employ $ \pi^3 $~\cite{wang2025pi} as the backbone for initial pose estimation, as it is permutation-equivariant and does not enforce the first frame as the origin of the coordinate system. Given a query image $I_q \in \mathbb{R}^{H \times W \times 3}$ and a set of retrieved reference images $\{I_{r,i}\}_{i=1}^K$, the network jointly processes these inputs to generate dense geometry predictions. Formally, let $\Phi_{\pi^3}$ denote the network function and $[\cdot]$ denote the concatenation along the sequence dimension. The feed-forward process is defined as:
\begin{equation}
    \mathcal{P}^{local}, \mathbf{P}^{local}, \mathcal{C} = \Phi_{\pi^3} \left( [I_q, I_{r,1}, \dots, I_{r,K}] \right)
    \label{eq:coarse_network}
\end{equation}
The output includes dense local point clouds $\mathcal{P}^{local} \in \mathbb{R}^{B \times H \times W \times 3}$ and a set of camera poses $\mathbf{P}^{local} \in SE(3)$ for both the query and reference images in a canonical coordinate system, where $B = K + 1$. In addition, for all input images, the network predicts confidence score maps $\mathcal{C}^{B \times H \times W}$ to measure the uncertainty of the prediction at each pixel. We select the reference image with the highest total confidence score as the anchor $I_{r,\text{anc}}$ for the subsequent stage.

\begin{figure*}[t!]
  \centering
  \includegraphics[width=\textwidth]{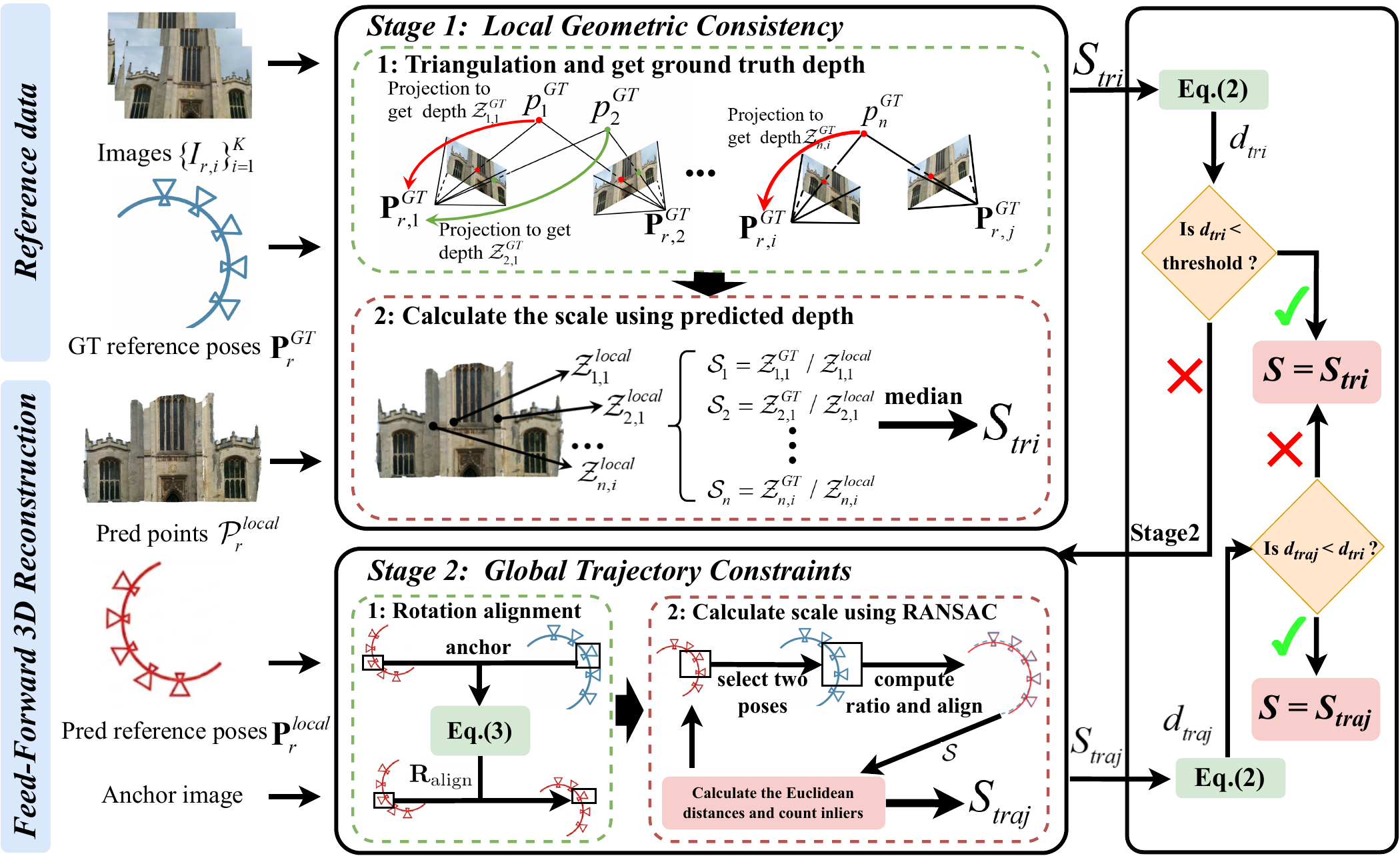} 
  \caption{\textbf{Scale Estimation Strategy. Stage 1:} Absolute depths are triangulated using ground truth (GT) reference poses. If the median ratio $S_{tri}$ between GT and local depths (from $\mathcal{P}_r^{local}$) has a deviation $d_{tri}$ below a threshold, $S_{tri}$ is adopted. \textbf{Stage 2:} Otherwise, we align the \textbf{\textcolor{red}{predicted trajectory $\textbf{P}^{local}_{r}$}} with the \textbf{\textcolor{blue}{GT trajectory $\textbf{P}^{GT}_{r}$}} via rotation $R_{\text{align}}$. A RANSAC scheme then yields the scale $S_{traj}$ by minimizing Euclidean distance error. $S_{traj}$ is accepted if $d_{traj}$ \textless $d_{tri}$ (fallback to $S_{tri}$ otherwise).}
  \label{fig:scale_estimation}
\end{figure*}

\noindent\textbf{Scale Estimation.}
\label{sec:scale_estimation} Since the network predictions lack an absolute scale $S$, we propose a two-stage scale estimation strategy that prioritizes local geometric consistency while enforcing global trajectory constraints as shown in \cref{fig:scale_estimation}.

 $\bullet$ \textbf{\textit{Stage 1: Local Geometric Consistency.}} We leverage the ground truth poses of retrieved reference images to estimate scale via triangulation. For geometric stability, we first sample image pairs $\{(I_{r,i}, I_{r,j})\}$ with a camera baseline empirically constrained within $[0.3, 10]$ meters. For each valid pair, we extract and match keypoints using SuperPoint~\cite{detone2018superpoint} and LightGlue~\cite{lindenberger2023lightglue}. These correspondences are then triangulated using the ground truth reference poses to yield a set of 3D points $\mathcal{P}^{GT}$.  Subsequently, each triangulated 3D point $p_n \in \mathcal{P}^{GT}$ associated with the image pair $(I_{r,i}, I_{r,j})$ is projected into the coordinate system of camera $i$ to obtain its absolute depth $\mathcal{Z}^{GT}_{n,i}$. Since the network predicts pixel-wise 3D coordinates in the local frame, we can extract the corresponding local depth $\mathcal{Z}^{local}_{n,i}$ by indexing the predicted local point cloud $\mathcal{P}_{r,i}^{local}$ of camera $i$ at the exact 2D pixel location of $p_n$. The final metric scale factor $S_{tri}$ is determined by taking the median of the ratios between the absolute and local depths.

 $\bullet$ \textbf{\textit{Stage 2: Global Trajectory Constraints.}} To prevent scale estimation failures in sparse views, we introduce a global consistency check. We formally define the trajectory radius $r$ as the root mean square distance from the $K$ reference camera centers $\{\mathbf{C}_i\}_{i=1}^K$ to their centroid $\bar{\mathbf{C}}$. To evaluate an estimated scale $S$, the relative deviation $d$ between the scaled local radius $r^{local}$ and the ground truth radius $r^{GT}$ is computed as:
        \begin{equation}
        \label{equ:r_d}
            r = \sqrt{\frac{1}{K} \sum_{i=1}^K \|\mathbf{C}_i - \bar{\mathbf{C}}\|_2^2}, \quad
            d = \left| \frac{S \cdot r^{local}}{r^{GT}} - 1 \right|
        \end{equation}
 Applying the triangulated scale $S_{tri}$ to the predicted reference poses yields an initial deviation $d_{tri}$. If $d_{tri}$ falls below a 5\% threshold, we adopt $S_{tri}$ as the final scale. Otherwise, we proceed to Stage 2, where we compute an alignment rotation matrix $\mathbf{R}_{\text{align}}$ using the anchor image $I_{r,\text{anc}}$, leveraging the scale-invariance of rotational transformations. Using its predicted rotation $\mathbf{R}_{r, \text{anc}}^{local}$ from $\mathbf{P}_{r, \text{anc}}^{local}$ and ground truth rotation $\mathbf{R}_{r, \text{anc}}^{GT}$ from $\mathbf{P}_{r, \text{anc}}^{GT}$, $\mathbf{R}_{\text{align}}$ is computed as:
    \begin{equation}
        \mathbf{R}_{\text{align}} = \mathbf{R}_{r, \text{anc}}^{GT} (\mathbf{R}_{r, \text{anc}}^{local})^\top
        \label{eq:r_align}
    \end{equation}
 Applying this matrix $\mathbf{R}_{\text{align}}$ aligns the orientation of the local coordinate system with the global coordinate system. After rotation alignment, we estimate the scale within a RANSAC~\cite{fischler1981random} framework. At each iteration, we randomly sample a pair of cameras and compute a candidate scale as the ratio of their ground truth Euclidean distance to the predicted distance. By applying this scale to the entire trajectory, we then compute the offset between the centroids of the predicted and ground truth trajectories as the global translation. We define cameras with a pose alignment error of $\leq$ 10 cm as inliers. After 500 iterations, the scale that maximizes the number of inliers is selected as $S_{traj}$. We then compare its corresponding trajectory deviation $d_{traj}$ against $d_{tri}$, ultimately adopting the scale associated with the smaller deviation.

\noindent\textbf{Pose initialization.}
\label{sec:pose_initialization}
Since all image poses are predicted in the local coordinate system, we compute the relative pose from the query to the anchor image. Utilizing the estimated metric scale $S$, the alignment rotation matrix $\mathbf{R}_{\text{align}}$, and the anchor's ground truth pose $\mathbf{P}_{r, \text{anc}}^{GT} = [\mathbf{R}_{r, \text{anc}}^{GT} | \mathbf{t}_{r, \text{anc}}^{GT}]$, the initial global pose of the query image $\mathbf{P}_{q}^{init} = [\mathbf{R}_{q}^{init} | \mathbf{t}_{q}^{init}]$ is derived as follows:
\begin{equation}
  \label{initial_pose_compute}
  \begin{aligned}
    \mathbf{R}_{q}^{\text{init}} &= \mathbf{R}_{\text{align}} \mathbf{R}_{q}^{\text{local}}, \\
    \mathbf{t}_{q}^{\text{init}} &= \mathbf{t}_{r, \text{anc}}^{GT} + \mathbf{R}_{\text{align}} \left( S \cdot (\mathbf{t}_{q}^{\text{local}} - \mathbf{t}_{r, \text{anc}}^{\text{local}}) \right)
  \end{aligned}
\end{equation}

\subsection{Pose Refinement}
\label{sec:pose_refinement}
With the predicted point cloud $\mathcal{P}_{r,i}^{\text{local}}$ and initial pose $\mathbf{P}_{q}^{init}$, we proceed to refine the pose. This process involves the following steps:

\noindent\textbf{Structure Optimization.} 
Unlike traditional SfM which relies on triangulation for initialization, we fully leverage the dense geometry provided by the network. This enables the efficient and stable acquisition of 3D structures, even under sparse view configurations. To ensure high-quality 2D-3D correspondences, we re-select the anchor image specifically for this optimization stage. We designate the first reference image as the new anchor due to its highest feature similarity with the query image, as validated in the supplementary material. We then construct global feature tracks by merging pairwise matches between this anchor and the other reference images. For each track, we initialize its 3D coordinates $\mathbf{X}_k$ by back-projecting the anchor's 2D keypoint using its corresponding scaled network-predicted depth. We then perform a structure-only Bundle Adjustment. By fixing the ground truth reference poses $\{\mathbf{P}_{r,i}^{GT}\}$ and refining only the 3D coordinates $\mathbf{X}_k$, we minimize the multi-view reprojection error:
\begin{equation}
    \arg\min_{\{\mathbf{X}_k\}} \sum_{k} \sum_{i \in \mathcal{O}_k} \rho \left( \|\pi(\mathbf{P}_{r,i}^{GT}, \mathbf{X}_k) - \mathbf{u}_{k, i}\|^2 \right)
\end{equation}
where $\pi(\cdot)$ is the projection function, $\mathcal{O}_k$ denotes the set of view indices (including the anchor) observing the $k$-th point, $\mathbf{u}_{k,i}$ is the observed 2D keypoint of track $k$ in view $i$, and $\rho(\cdot)$ is the Soft $L1$ loss function.

\noindent\textbf{Matching.} 
We project the optimized 3D points $\{\mathbf{X}_k\}$ onto the query image:
\begin{equation}
    \hat{\mathbf{u}}_{k} = \pi(\mathbf{P}_{q}^{init}, \mathbf{X}_k)
\end{equation}
For each projected point $\hat{\mathbf{u}}_{k}$, we define a local search region (set to a 20-pixel radius in our experiments) and retrieve candidate keypoints extracted from the query image. We compute the descriptor distance between the reference descriptors associated with the 3D points and the candidate query descriptors. Matches are established using a nearest neighbor search within the local region, effectively filtering out outliers and reducing search complexity.

\begin{table*}[t]
  \caption{\textbf{Results on 7Scenes dataset.} We compare the median translation and rotation errors (cm/$^{\circ}$) of different methods, lower is better. Best results are in \textbf{bold}.
  }
  \label{tab:7scenes_dense_view}
  \centering
  \resizebox{\textwidth}{!}{
  \begin{tabular}{lccccccccc}
    \toprule
    & Methods & Chess & Fire & Heads & Office & Pumpkin & Redkitchen & Stairs & Avg. $\downarrow$[cm/$^{\circ}$] \\
    \midrule
    \multirow{3}{*}{APR}
        & PoseNet \cite{kendall2015posenet} \footnotemark[1] & 10/4.02 & 27/10.0 & 18/13.0 & 17/5.97 & 19/4.67 & 22/5.91 & 35/10.5 & 21/7.74 \\
        & DFNet \cite{chen2022dfnet} \footnotemark[1] & 3/1.12 & 6/2.30 & 4/2.29 & 6/1.54 & 7/1.92 & 7/1.74 & 12/2.63 & 6/1.93 \\
        & Marepo \cite{chen2024map} \footnotemark[2] & 1.9/0.83 & 2.3/0.92 & 2.1/1.24 & 2.9/0.93 & 2.5/0.88 & 2.9/0.98 & 5.9/1.48 & 2.9/1.04 \\
    \midrule
    \multirow{3}{*}{SCR}
        & DSAC* \cite{brachmann2021visual} \footnotemark[2] & \textbf{0.5}/0.17 & 0.8/0.28 & 0.5/0.34 & 1.2/0.34 & 1.2/0.28 & \textbf{0.7}/0.21 & 2.7/0.78 & 1.1/0.34 \\
        & ACE \cite{brachmann2023accelerated} & \textbf{0.5}/0.19 & 0.8/0.33 & 0.5/0.33 & 1.1/0.29 & 1.1/\textbf{0.22} & 0.8/0.20 & 2.9/0.80 & 1.1/0.34 \\
        & GLACE \cite{wang2024glace} & 0.6/0.18 & 0.9/0.34 & 0.6/0.35 & 1.0/0.29 & \textbf{0.9}/0.23 & 0.8/0.20 & 3.1/0.95 & 1.1/0.36\\
    \midrule
    \multirow{6}{*}{NVS}
        & pNeRFLoc \cite{zhao2024pnerfloc}  & 2/0.8 & 2/0.88 & 1/0.83 & 3/1.05 & 6/1.51 & 5/1.54 & 32/5.73 & 7.3/1.76 \\
        & DFNet + $\text{NeFeS}_{50}$ \cite{chen2024neural} & 2/0.57 & 2/0.74 & 2/1.28 & 2/0.56 & 2/0.55 & 2/0.57 & 5/1.28 & 2.4/0.79 \\
        & HR-APR \cite{liu2024hr} & 2/0.55 & 2/0.75 & 2/1.45 & 2/0.64 & 2/0.62 & 2/0.67 & 5/1.30 & 2.4/0.85  \\
        & NeRFMatch \cite{zhou2024nerfect} & 0.9/0.3 & 1.1/0.4 & 1.5/1.0 & 3.0/0.8 & 2.2/0.6 & 1.0/0.3 & 10.1/1.7 & 2.8/0.7  \\
        & MCLoc \cite{trivigno2024unreasonable} & 2/0.8 & 3/1.4 & 3/1.3 & 4/1.3 & 5/1.6 & 6/1.6 & 6/2.0 & 4.1/1.43 \\
        & ACE + GS-CPR \cite{liugs} & \textbf{0.5/0.15} & \textbf{0.6/0.25} & \textbf{0.4/0.28} & \textbf{0.9/0.26} & 1.0/0.23 & \textbf{0.7/0.17} & \textbf{1.4/0.42} & \textbf{0.8/0.25}\\
    \midrule
        & \textcolor{blue}{$L^3$} (ours) & 0.6/0.21 & 1.2/0.46 & 0.7/0.41 & 1.1/0.32 & 1.4/0.34 & 1.0/0.26 & 3.1/0.87 & 1.3/0.41  \\
    \bottomrule
  \end{tabular}
  }
\end{table*}

\begin{table*}[!ht]
    \centering
    \caption{\textbf{Results on 12Scenes dataset.} We compare the median translation and rotation errors (cm/$^{\circ}$) of different methods, and the average accuracy of frames under a specified threshold. Best results are in \textbf{bold}.}
    \label{tab:12scenes_dense_view}
    \resizebox{0.8\textwidth}{!}{
    \begin{tabular}{lcccc}
        \toprule
        & Methods  & Avg. $\downarrow$[cm/$^{\circ}$] & Avg. $\uparrow$[5cm, 5$^\circ$] & Avg. $\uparrow$[1cm, 1$^\circ$] \\
        \midrule
        \multirow{1}{*}{APR}
            & Marepo \cite{chen2024map} \footnotemark[2] & 2.1/1.04 & 95 & $-$ \\
        \midrule
        \multirow{2}{*}{SCR}
            & ACE \cite{brachmann2023accelerated} & 0.7/0.26 & \textbf{100} & 75.1 \\
            & GLACE \cite{wang2024glace} & 0.7/0.25 & \textbf{100} & 78.7 \\
        \midrule
        \multirow{1}{*}{NVS}
            & ACE + GS-CPR \cite{liugs} & 0.5/0.21 & \textbf{100} & \textbf{86.7} \\
        \midrule
            & \textcolor{blue}{$L^3$} (ours) & \textbf{0.4/0.19} & 99.5 & \textbf{86.7} \\
        \bottomrule
    \end{tabular}
    }
\end{table*}

\noindent\textbf{PnP Refinement.}
The guided matching process yields a set of 2D-3D correspondences $\mathcal{M} = \{(\mathbf{u}_{q,k}, \mathbf{X}_k)\}$. We estimate the final 6-DoF query pose $\mathbf{P}_{q}$ by solving the PnP problem. Taking the coarse pose $\mathbf{P}_{q}^{init}$ as the initialization, we apply RANSAC~\cite{fischler1981random} and then refine the pose on the inlier set by using the Levenberg-Marquardt algorithm:
\begin{equation}
    \mathbf{P}_{q}^{*} = \arg\min_{\mathbf{P}} \sum_{(\mathbf{u}, \mathbf{X}) \in \mathcal{M}} \left\| \pi(\mathbf{P}, \mathbf{X}) - \mathbf{u} \right\|^2
\end{equation}
To ensure maximum robustness, we evaluate both the refined pose $\mathbf{P}_{q}^{*}$ and the initial pose $\mathbf{P}_{q}^{init}$ based on their inlier counts with respect to the query image. The pose yielding the larger inlier count is ultimately selected as the final result.

\section{Experiments}
\label{sec:experiments}
\label{sec:experiment setup}
\noindent$\bullet$ \textbf{Datasets.} We evaluate \textcolor{blue}{$L^3$} on three benchmarks: the indoor \textbf{\textit{7Scenes}}~\cite{shotton2013scene} and \textbf{\textit{12Scenes}}~\cite{valentin2016learning} datasets, utilizing SfM ground truth from~\cite{brachmann2021limits}, and the outdoor \textbf{\textit{Cambridge Landmarks}}~\cite{kendall2015posenet}. While the indoor sets provide varied architectural scales, the outdoor scenes introduce challenging dynamic objects and lighting variations to assess robustness "in the wild".
\footnotetext[1]{These data are cited from NeFeS~\cite{chen2024neural}.}
\footnotetext[2]{These data are cited from GS-CPR~\cite{liugs}.}
\noindent$\bullet$ \textbf{Metrics.} We assess performance using median \textbf{\textit{translation}} ($cm$) and \textbf{\textit{rotation}} ($^{\circ}$) errors w.r.t camera pose, alongside the \textbf{\textit{recall rate}} to measure the percentage of successful localized rate within specified error thresholds.

\noindent $\bullet$ \textbf{Setup.} Reference images are retrieved via MegaLoc~\cite{berton2025megaloc} and filtered with a 0.3m minimum physical baseline to ensure spatial diversity. By default, $K=10$ reference images are selected per query. All experiments were conducted on an NVIDIA H20 GPU.

\noindent$\bullet$ \textbf{Baselines.} We compare \textcolor{blue}{$L^3$} against representative image-based APR methods, including PoseNet~\cite{kendall2015posenet}, DFNet~\cite{chen2022dfnet}, and Marepo~\cite{chen2024map}. Structure-based baselines encompass SCR networks (DSAC*~\cite{brachmann2021visual}, ACE~\cite{brachmann2023accelerated}, GLACE~\cite{wang2024glace}) and recent NVS approaches, including NeRF-based variants~\cite{liu2024hr,zhao2024pnerfloc,zhou2024nerfect,chen2024neural}, the representation-agnostic MCLoc~\cite{trivigno2024unreasonable}, and the 3DGS-based GS-CPR~\cite{liugs} (initialized with ACE). For outdoor evaluation on Cambridge Landmarks, we further include feature matching-based Active Search~\cite{sattler2016efficient}, HLoc~\cite{SarlinCSD19}, and ImLoc~\cite{jiang2026imloc}.

\subsection{Dense View Localization}
\label{sec:dense_view_localization}
We first evaluate \textcolor{blue}{$L^3$} under dense-view configurations, utilizing the complete set of reference images in each dataset. 

\noindent$\bullet$ \textbf{Indoor Localization.} Evaluation results on the 7Scenes dataset are summarized in \cref{tab:7scenes_dense_view}. Without requiring scene-specific training or reconstruction, \textcolor{blue}{$L^3$} significantly outperforms APR methods (e.g., DFNet, Marepo) and NeRF-based synthesis approaches (e.g., pNeRFLoc, NeRFMatch), while remaining competitive with leading SCR networks like ACE and GLACE. On the more extensive 12Scenes dataset (\cref{tab:12scenes_dense_view}), \textcolor{blue}{$L^3$} establishes a new SOTA, surpassing established SCR frameworks—including ACE and GLACE—as well as the recent NVS-based GS-CPR. These results underscore the efficacy of our scene-agnostic paradigm, which matches or exceeds the precision of structure-based methods without the burden of offline pre-processing.

\begin{table*}[t!]
    \caption{\textbf{Results on Cambridge Landmarks dataset}. We compare the median translation and rotation errors (cm/$^{\circ}$) of different methods. Best results are in \textbf{bold}.}
    \label{tab:cambridge_dense_view}
    \centering
    \resizebox{0.9\textwidth}{!}{
    \begin{tabular}{lccccccc}
        \toprule
        & Methods & Kings & Hospital & Shop & Church & Avg. $\downarrow$[cm/$^{\circ}$] \\
        \midrule
        \multirow{3}{*}{FM}
            & AS (SIFT) \cite{sattler2016efficient} & 13/\textbf{0.2} &  20/0.4 & \textbf{4/0.2} & 8/0.3 & 11/0.28 \\
            & HLoc (SP+SG) \cite{detone2018superpoint,sarlin2020superglue} & 12/\textbf{0.2} & 15/\textbf{0.3} & \textbf{4/0.2} & \textbf{7/0.2} & 10/0.23 \\
            & ImLoc \cite{jiang2026imloc} & \textbf{11/0.2} & \textbf{14/0.3} & \textbf{4/0.2} & \textbf{7/0.2} & \textbf{9/0.23} \\
        \midrule
        \multirow{3}{*}{APR}
            & LENS \cite{moreau2022lens} & 33/0.5 & 44/0.9 & 27/1.6 & 53/1.6  & 39/1.15\\
            & DFNet \cite{chen2022dfnet} & 73/2.37 & 200/2.98 & 67/2.21 & 137/4.03 & 119/2.90 \\
            & PMNet \cite{lin2024learning} & 68/1.97 & 103/1.31 & 58/2.10 & 133/3.73 & 90/2.27 \\
        \midrule
        \multirow{3}{*}{SCR}
            & DSAC* \cite{brachmann2021visual} & 15/0.3 & 21/0.4 & 5/0.3 & 13/0.4 & 14/0.35 \\
            & ACE \cite{brachmann2023accelerated} & 28/0.37 & 32/0.63 & 5/0.28 & 19/0.60 & 21/0.47 \\
            & GLACE \cite{wang2024glace} & 18/0.32 & 19/0.42 & 5/0.22 & 9/0.29 & 13/0.31 \\
        \midrule
        \multirow{5}{*}{NVS}
            & CrossFire \cite{moreau2023crossfire} & 47/0.7 & 43/0.7 & 20/1.2 & 39/1.4 & 37/1 \\
            & DFNet + $\text{NeFeS}_{50}$ \cite{chen2024neural} & 37/0.54 & 52/0.88 & 15/0.53 & 37/1.14 & 35/0.77 \\
            & HR-APR \cite{liu2024hr} & 36/0.58 & 53/0.89 & 13/0.51 & 38/1.16 & 35/0.78 \\
            & MCLoc \cite{trivigno2024unreasonable} & 31/0.42 & 39/0.73 & 12/0.45 & 26/0.88 & 27/0.62 \\
            & ACE + GS-CPR \cite{liugs} & 20/0.29 & 21/0.40 & 5/0.24 & 13/0.40 & 15/0.33 \\
        \midrule
            & \textcolor{blue}{$L^3$} (ours) & 13/0.22 & 15/\textbf{0.30} & 5/0.23 & 9/0.31 & 11/0.27 \\
         \bottomrule
    \end{tabular}
    }
\end{table*}

\noindent$\bullet$ \textbf{Outdoor Localization.} As reported in \cref{tab:cambridge_dense_view}, \textcolor{blue}{$L^3$} delivers performance competitive with leading FM pipelines like HLoc and ImLoc while significantly outperforming other baselines. Notably, despite the vastly increased scene scale, \textcolor{blue}{$L^3$} requires no hyperparameter tuning or dataset-specific modifications, underscoring its superior generalization over existing scene-specific solutions.

\subsection{Sparse Scene Localization}
\label{sec:sparse_view_localization}
\begin{figure*}[t!]
  \centering
  \begin{subfigure}[b]{0.95\linewidth}
    \centering
    \includegraphics[width=\linewidth]{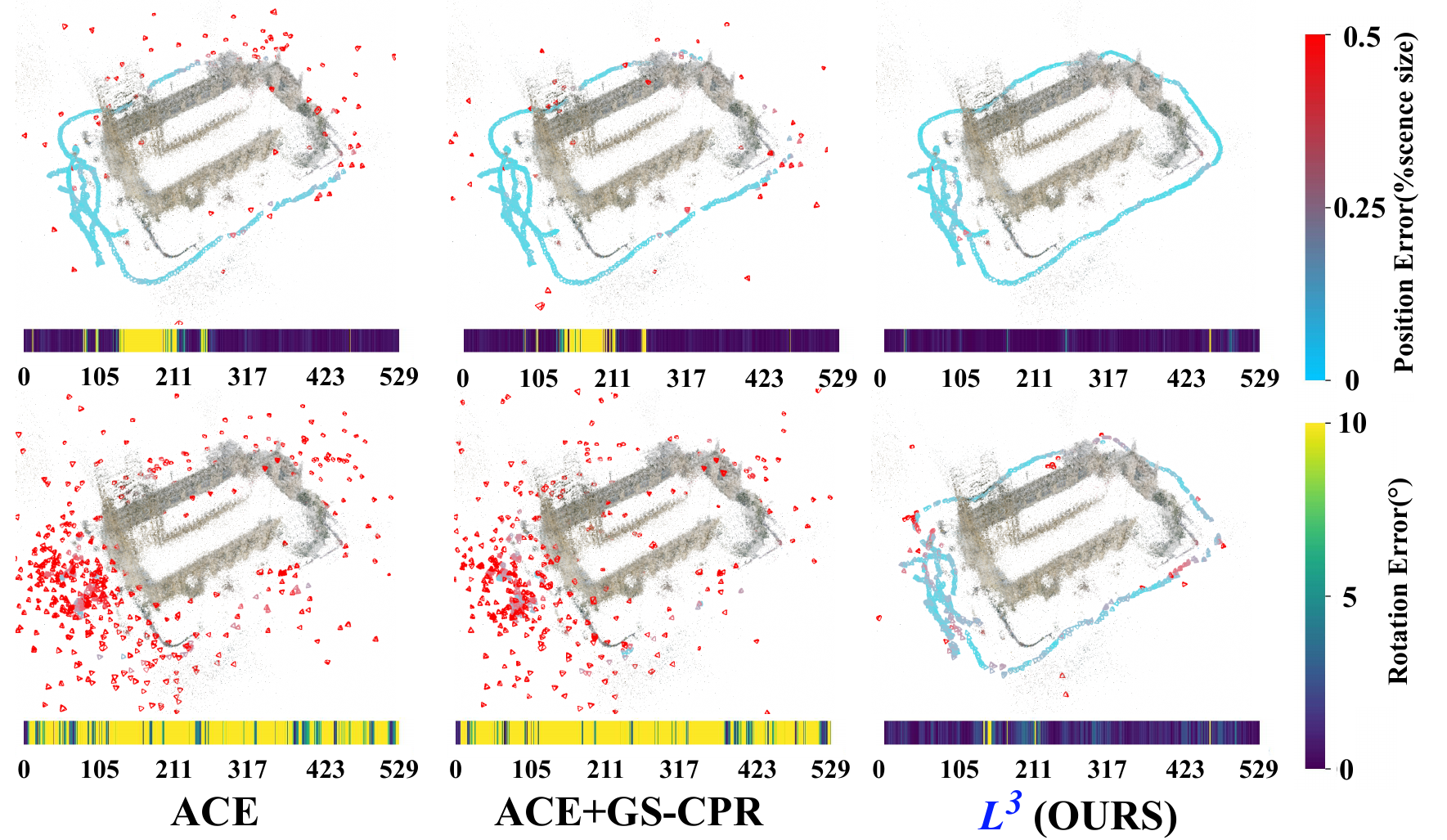} 
    \caption{StMarysChurch}
    \label{fig:cambridge}
  \end{subfigure}
  \begin{subfigure}[b]{0.94\linewidth}
    \centering
    \includegraphics[width=\linewidth]{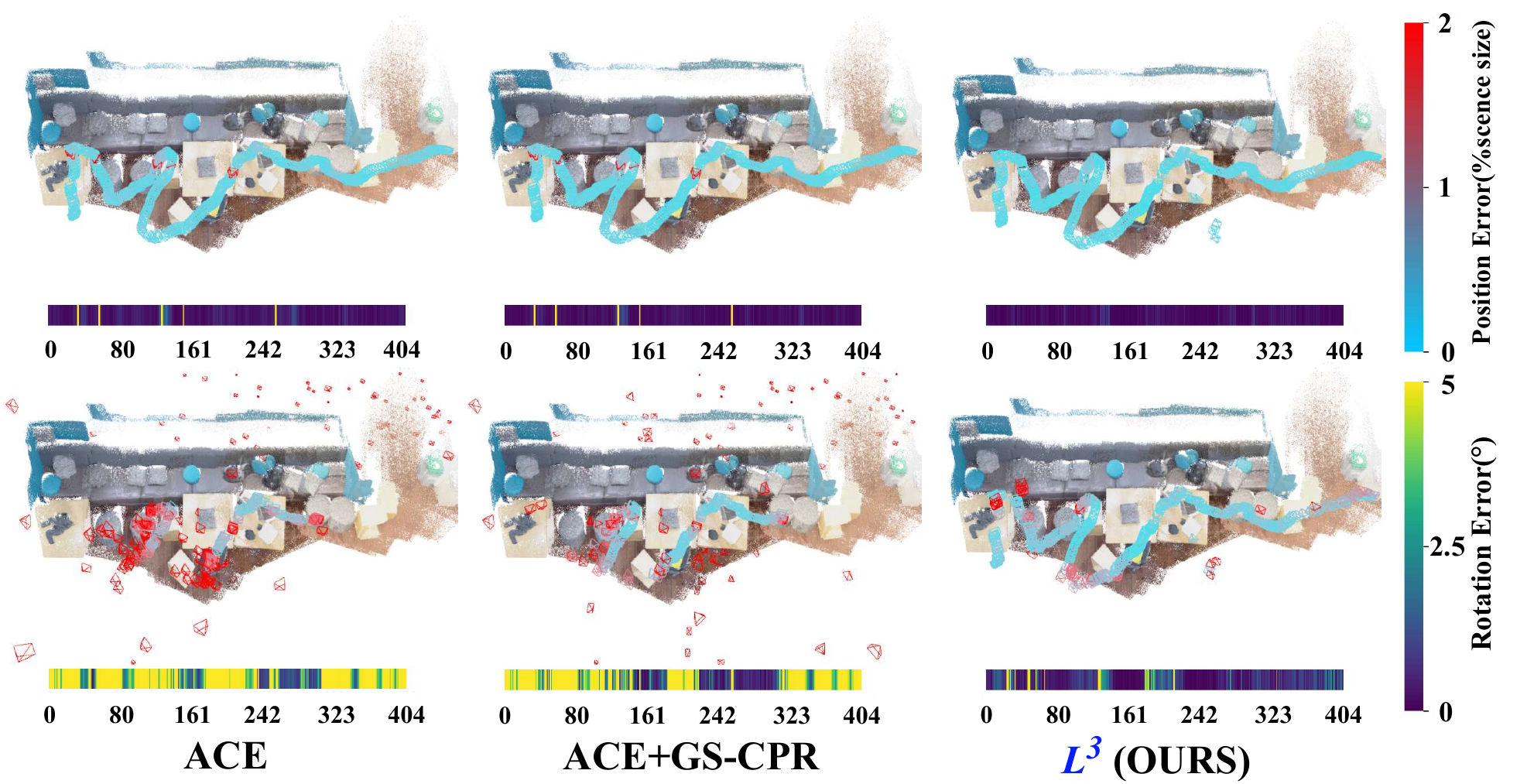}
    \caption{Office2\_5b}
    \label{fig:12scenes}
  \end{subfigure}
  \caption{\textbf{Comparison of localization errors.} Position (trajectories) and rotation (bottom bars, x-axis denotes frame indices) errors are colored by magnitude for ACE~\cite{brachmann2023accelerated}, GS-CPR~\cite{liugs}, and our \textcolor{blue}{$L^3$}. (a) \textbf{Outdoor StMarysChurch}  (Cambridge Landmarks). (b) \textbf{Indoor Office2\_5b}  (12Scenes). Both panels compare \textbf{dense (top)} and \textbf{sparse (bottom, $N=30$)} settings.}
  \label{fig:trajectories_comparison}
\end{figure*}

To evaluate performance under data scarcity, reference image sets are uniformly downsampled into sparse subsets. Accounting for variations in scene scale, we retain $N \in [5, 30]$ images per scene for indoor datasets (7Scenes and 12Scenes) and $N \in [30, 60]$ for the outdoor Cambridge Landmarks. In the extreme $N=5$ indoor configuration, each query is restricted to retrieving only these five available reference images. We visualize the estimated trajectories and localization errors across different methods under dense and sparse conditions in \cref{fig:trajectories_comparison}. As depicted, our method exhibits superior stability compared to baselines.

\begin{figure*}[t!]
  \centering
  \begin{subfigure}[b]{0.33\linewidth}
    \centering
    \includegraphics[width=\linewidth]{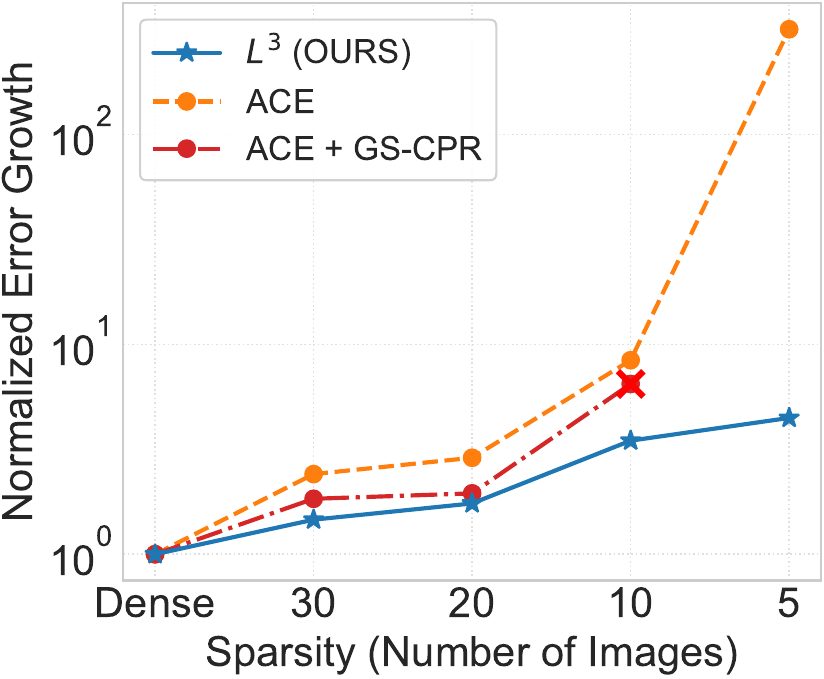} 
    \caption{7Scenes}
    \label{fig:7scenes_sparse_robustness}
  \end{subfigure}
  \hfill
  \begin{subfigure}[b]{0.33\linewidth}
    \centering
    \includegraphics[width=\linewidth]{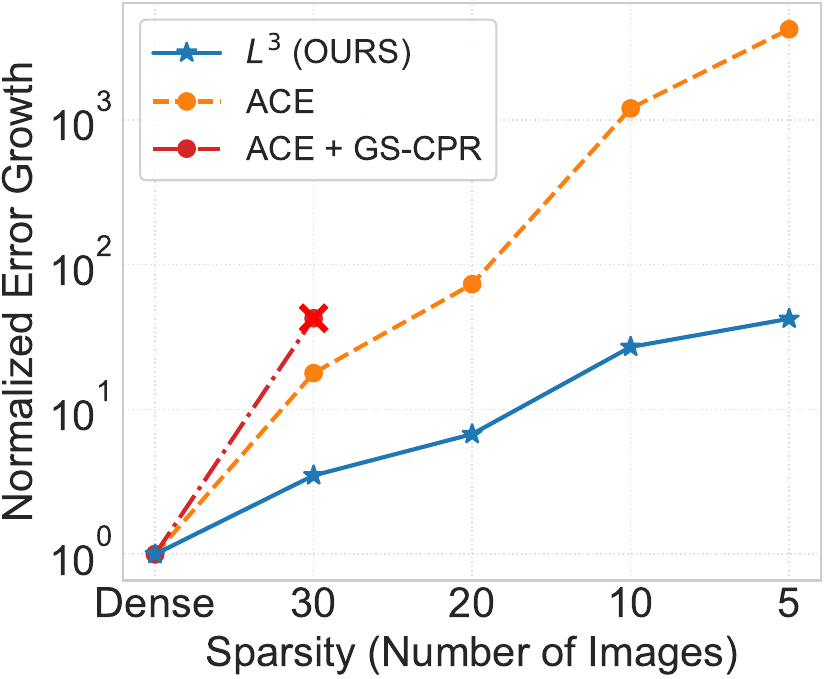}
    \caption{12Scenes}
    \label{fig:12scenes_sparse_robustness}
  \end{subfigure}
  \hfill
  \begin{subfigure}[b]{0.33\linewidth}
    \centering
    \includegraphics[width=\linewidth]{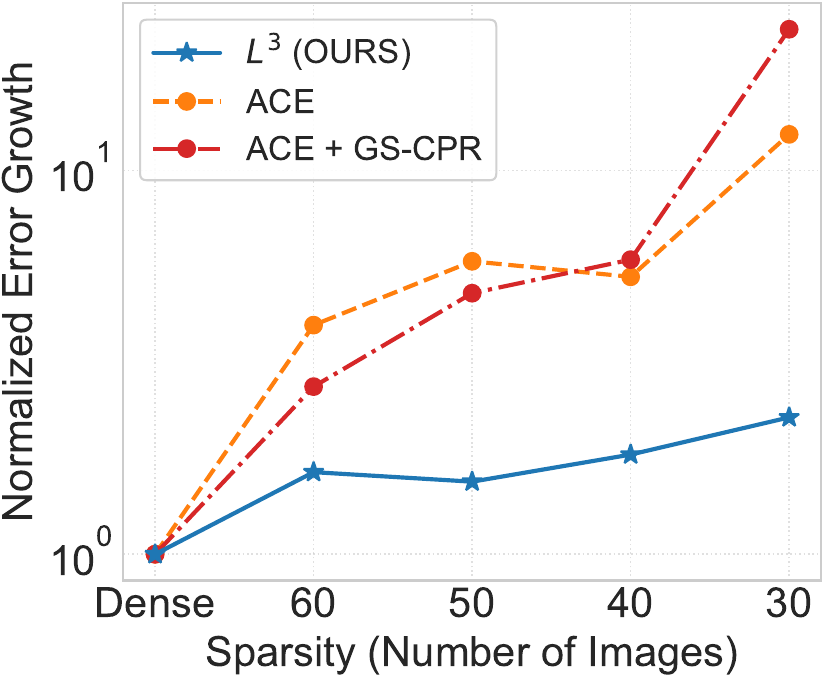}
    \caption{Cambridge Landmarks}
    \label{fig:cambridge_sparse_robustness}
  \end{subfigure}
  \caption{\textbf{Robustness analysis under increasing sparsity.} We plot the log-scale translation error growth relative to the dense setting on three datasets. "\textcolor{red}{\XSolid}" indicates localization failure beyond that sparsity level.}
  \label{fig:sparse_growth}
\end{figure*}

As illustrated in \cref{fig:sparse_growth}, \textcolor{blue}{$L^3$} exhibits superior robustness by maintaining the lowest and most stable error growth curves across all benchmarks as sparsity increases. Quantitative results in \cref{tab:7scenes_12scenes_sparse_views} reveal that ACE is highly sensitive to data scarcity, often diverging in sparse settings. While GS-CPR performs well on 7Scenes with $N \ge 20$, it fails when reference images are insufficient to support 3DGS training, such as on 12Scenes or the extreme $N=5$ subset. In contrast, \textcolor{blue}{$L^3$} establishes SOTA performance across all 12Scenes configurations and remains the only framework to maintain stable localization at $N=5$ on both indoor datasets. On Cambridge Landmarks (\cref{tab:cambridge_sparse_views}), where ACE and GS-CPR accuracy drops sharply, \textcolor{blue}{$L^3$} consistently delivers high precision. Notably, our performance at $N=30$ is comparable to ACE’s results in the fully dense setting and outpaces GS-CPR by more than an order of magnitude.

\begin{table*}[t]
    \caption{\textbf{Sparse view localization on the 7Scenes and 12Scenes datasets.} We report the mean of median translation and rotation errors (cm/$^{\circ}$) across all subscenes. $N$ is the number of sampled reference images. "-" indicates localization failure in at least one subscene. Best results are in \textbf{bold}.}
    \label{tab:7scenes_12scenes_sparse_views}
    \centering
    \resizebox{0.9\textwidth}{!}{
    \begin{tabular}{ccccccc}
        \toprule
        \multirow{2}{*}{Datasets} & \multirow{2}{*}{Methods} & \multirow{2}{*}{Dense} & \multicolumn{4}{c}{Sparse} \\
        \cmidrule(lr){4-7}
        & & & N=30 & N=20 & N=10 & N=5 \\
        \midrule
        \multirow{5}{*}{7Scenes}
            & ACE~\cite{brachmann2023accelerated}  & 1.1/0.34 & 2.7/0.65 & 3.2/0.79 & 9.2/1.53 & 348.7/4.78 \\
            & ACE + GS-CPR~\cite{liugs} & \textbf{0.8/0.25} & \textbf{1.5/0.42} & \textbf{1.6/0.46} & 5.2/1.33 & $-$ \\
            & \textcolor{blue}{${L^3}_\text{coarse}$} (ours) & 2.5/0.71 & 3.7/1.02 & 4.3/1.17 & 5.7/1.52 & 6.4/1.65 \\
            & \textcolor{blue}{$L^3_\text{w/o. Optim}$} (ours) & 2.5/0.67 & 2.8/0.79 & 3.1/0.84 & 5.1/1.42 & 6.5/1.66 \\
            & \textcolor{blue}{$L^3$} (ours) & 1.3/0.41 & 1.9/0.56 & 2.3/0.63 & \textbf{4.5/1.26} & \textbf{5.9/1.61} \\
        \midrule
        \multirow{5}{*}{12Scenes}
            & ACE~\cite{brachmann2023accelerated} & 0.7/0.26 & 12.5/1.32 & 51.3/3.76 & 837.9/12.1 & 2950/20.63 \\
            & ACE + GS-CPR~\cite{liugs} & 0.5/0.21 & 21.3/0.95 & $-$ & $-$ & $-$ \\
            & \textcolor{blue}{${L^3}_\text{coarse}$} (ours) & 1.0/0.36 & 2.6/0.90 & 3.5/1.12 & 10.8/2.52 & \textbf{16.9/2.56} \\
            & \textcolor{blue}{$L^3_\text{w/o. Optim}$} (ours) & 0.7/0.25 & 1.6/0.54 & 2.8/0.94 & 10.7/2.53 & 17.2/2.57 \\
            & \textcolor{blue}{$L^3$} (ours) & \textbf{0.4/0.19} & \textbf{1.4/0.53} & \textbf{2.7/0.93} & \textbf{10.2/2.51} & \textbf{16.9/2.56} \\
        \bottomrule
    \end{tabular}
    }
\end{table*}

\begin{table*}[t!]
    \caption{\textbf{Results of sparse view localization on the Cambridge dataset.} We compare the mean of the median translation and rotation errors (cm/$^{\circ}$) for different methods across all subscenes. Best results are in \textbf{bold}.}
    \label{tab:cambridge_sparse_views}
    \centering
    \resizebox{0.88\textwidth}{!}{
    \begin{tabular}{cccccc}
        \toprule
            \multirow{2}{*}{Methods} & \multirow{2}{*}{Dense} & \multicolumn{4}{c}{Sparse} \\
            \cmidrule(lr){3-6}
            & &N=60 & N=50 & N=40 & N=30 \\
            \midrule
            ACE~\cite{brachmann2023accelerated}  & 21/0.47 & 83/1.70 & 122/2.85 & 111/2.43 & 261/5.65 \\
            ACE + GS-CPR~\cite{liugs} & 15/0.33  & 41/1.10 & 72/2.13 & 88/5.41 & 350/8.63 \\
            \textcolor{blue}{${L^3}_\text{coarse}$} (ours) & 36/0.54 & 42/0.64 & 40/0.64 & 56/0.68 & 54/0.76 \\
            \textcolor{blue}{$L^3_\text{w/o. Optim}$} (ours) & 15/0.33 & 23/0.46 & 20/0.53 & 23/0.47 & 34/0.66 \\
            \textcolor{blue}{$L^3$} (ours) & \textbf{11/0.27} &\textbf{18/0.39} & \textbf{17/0.37} & \textbf{20/0.42} & \textbf{25/0.53} \\
        \bottomrule
    \end{tabular}
    }
\end{table*}

\begin{figure*}[t!]
  \centering
  \begin{subfigure}[b]{0.43\linewidth}
    \centering
    \includegraphics[width=\linewidth]{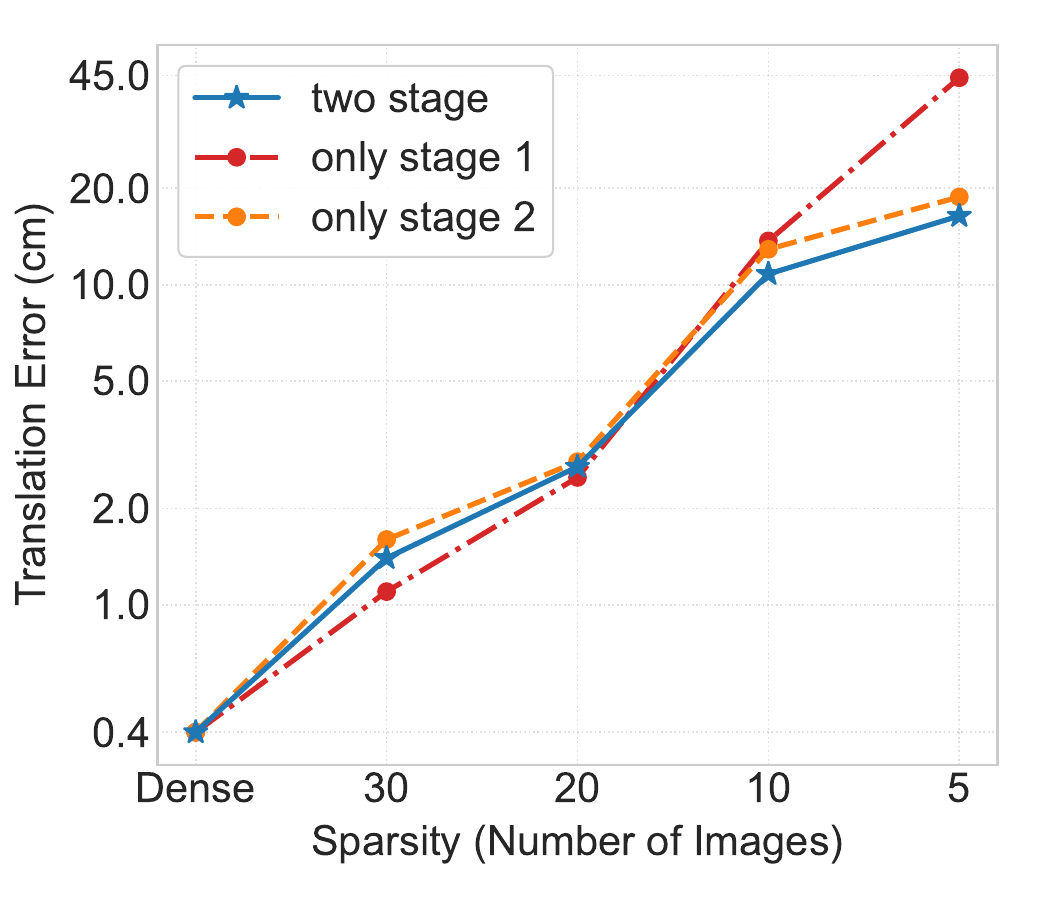} 
    \caption{translation error}
    \label{fig:translation_robustness}
  \end{subfigure}
  \begin{subfigure}[b]{0.43\linewidth}
    \centering
    \includegraphics[width=\linewidth]{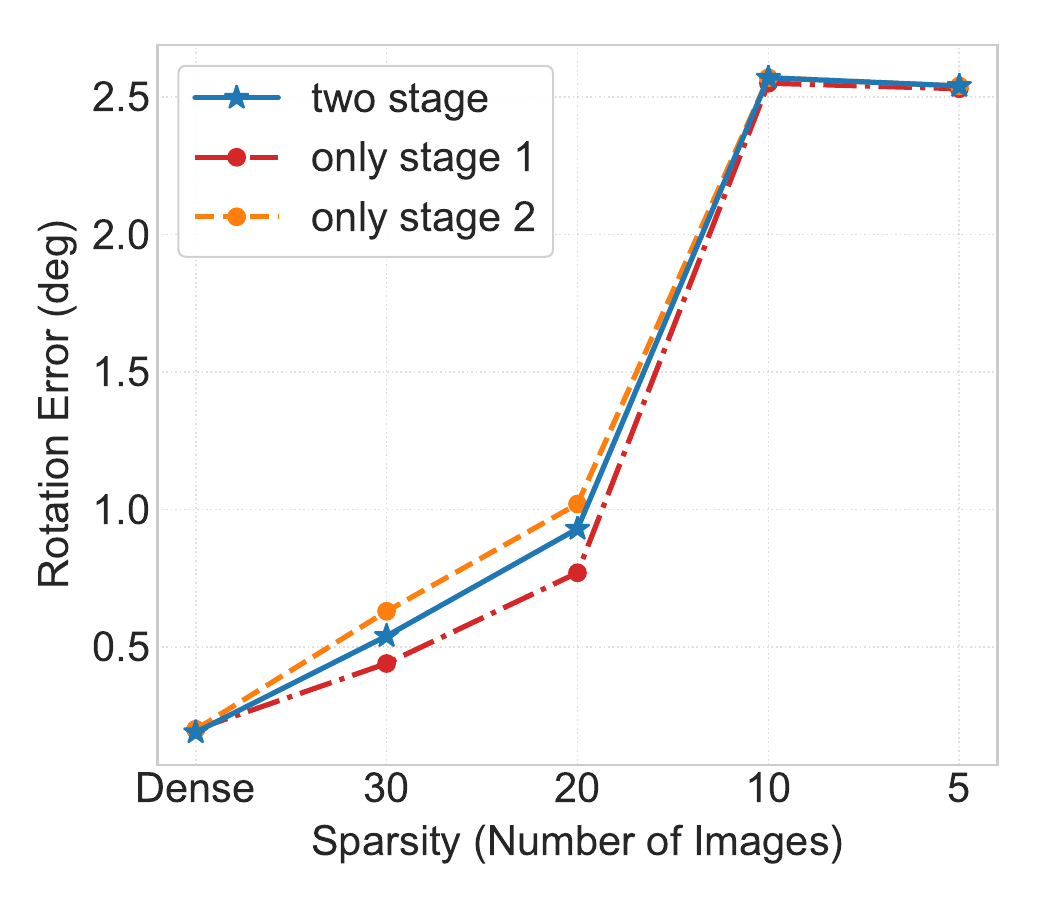}
    \caption{rotation error}
    \label{fig:rotation_robustness}
  \end{subfigure}
  \caption{\textbf{Ablation of scale estimation strategies on 12Scenes dataset.} We compare our full method in \cref{sec:scale_estimation} (\textbf{blue line}) against variants using only stage 1 (local geometric consistency, \textbf{red line}) or stage 2 (global trajectory constraints, \textbf{yellow line}).}
  \label{fig:ablation_scale_recovery_strategy}
\end{figure*}

\subsection{Ablation Study}
\label{sec:ablation_study}
We ablate scale estimation strategy and pose refinement one by one to quantitatively verify the effectiveness of these two vital modules in \textcolor{blue}{$L^3$}. 

\noindent$\bullet$ \textbf{Analysis of Scale Estimation Strategy.} As illustrated in \cref{fig:ablation_scale_recovery_strategy}, our two scale recovery components exhibit distinct complementarity across varying data densities. While local geometric triangulation provides superior precision in dense scenarios, it suffers from unreliable results in sparse settings ($N \le 10$) due to the scarcity of effective matches. Conversely, RANSAC-based global trajectory alignment—though less precise in dense scenes—effectively leverages the reconstruction network's inherent robustness to avoid the drastic error spikes typical of triangulation under data scarcity. By synergizing these strengths, our complete two-stage strategy achieves optimal overall performance and ensures sustained robustness against fluctuations in data density.

\noindent$\bullet$ \textbf{Effectiveness of Pose Refinement.} We define \textcolor{blue}{$L^3_\text{coarse}$} as the direct network output without the refinement stage, and \textcolor{blue}{$L^3_\text{w/o. Optim}$} as the output of PnP using network predictions without the structure optimization stage. As evidenced in \cref{tab:7scenes_12scenes_sparse_views} and \cref{tab:cambridge_sparse_views}, the pose refinement module significantly enhances localization accuracy across nearly all indoor and outdoor benchmarks. Furthermore, utilizing optimized 3D points for PnP further reduces the localization error. Notably, in sparse scenarios, while \textcolor{blue}{$L^3_\text{coarse}$} inherently demonstrates superior stability compared to existing methods, the refinement module further consistently minimizes localization errors. Crucially, in extremely sparse settings (e.g., $N=10$ in 12Scenes) where limited optimization points may yield unreliable PnP solutions, our module employs a query-image inlier constraint to retain the initial prediction selectively. This robust fallback mechanism effectively averts the performance degradation observed in methods like GS-CPR. 

\section{Discussion}
\label{sec:discussion}
\noindent$\bullet$ \textbf{What are the advantages of \textcolor{blue}{$L^3$} compared to previous methods?} \textcolor{blue}{$L^3$} establishes a ``zero-mapping'' paradigm for visual localization, achieving high accuracy without any scene-specific offline training or reconstruction. By leveraging the generalization of feed-forward networks such as $\pi^3$~\cite{wang2025pi}, our framework performs direct online inference, eliminating the hours of SfM or training required by traditional structure-based methods. This scene-agnostic approach proves exceptionally robust in sparse-view settings where baselines such as ACE~\cite{brachmann2023accelerated} and GS-CPR~\cite{liugs} often diverge. Our evaluations verify that the two-stage scale estimation strategy maintains metric consistency under data scarcity, while the query-image inlier check in our refinement module provides a vital fallback to prevent performance degradation. Ultimately, \textcolor{blue}{$L^3$} enables rapid, reliable deployment in uncharted environments with minimal reference data. 

\noindent$\bullet$ \textbf{What are the limitations of \textcolor{blue}{$L^3$}?} 
Despite its robust performance, \textcolor{blue}{$L^3$} is subject to several limitations. As shown in \cref{tab:time_analysis}, the primary bottleneck is inference latency, which averages 2.1 seconds per query. Additionally, high computational demands limit direct deployment on resource-constrained edge hardware.

\noindent$\bullet$ \textbf{Potential Applications.} By eliminating the prohibitive preprocessing and storage costs of traditional mapping, \textcolor{blue}{$L^3$} enables instant, scalable localization in uncharted environments using only sparse reference data. While current latency limits real-time edge execution, this scene-agnostic paradigm is ideal for latency-tolerant or distributed architectures. Promising applications include HD mapping for autonomous driving, cloud-edge pose initialization for VR/AR, and the immediate deployment of robotics in unknown terrains.

\section{Conclusion}
We present \textcolor{blue}{$L^3$}, a paradigm-shifting framework that validates the feasibility of replacing intensive offline preprocessing with direct online 3D reconstruction for visual localization. By leveraging feed-forward dense predictions alongside a robust two-stage scale recovery and pose refinement pipeline, \textcolor{blue}{$L^3$} achieves accuracy competitive with SOTAs without the need for pre-built maps or scene-specific optimization. Crucially, our approach exhibits exceptional robustness in sparse-view scenarios, significantly outperforming existing solutions under extreme data constraints. This work establishes a new 'zero-mapping' benchmark, facilitating rapid and reliable deployment in unconstrained environments while drastically reducing the storage and computational overhead for practical engineering applications.

\begin{table}[tb]
    \caption{\textbf{Time comparison on the Cambridge Landmarks dataset.} We only need top retrieval. Best results are in \textbf{bold}.}
    \label{tab:time_analysis}
    \centering
    \resizebox{0.5\textwidth}{!}{
    \begin{tabular}{lccc}
        \toprule
            Metrics & ACE~\cite{brachmann2023accelerated} & GS-CPR~\cite{liugs} & \textcolor{blue}{$L^3$} (ours) \\
        \midrule
            Pre-processing time (min) $\downarrow$ & 2 & 31 & \textbf{0.6} \\
            Inference time (s) $\downarrow$        & \textbf{0.02} & 0.26 & 2.1 \\
            Storage (MB) $\downarrow$              & 4 & 203 & \textbf{0} \\
            Avg. Error [cm/$^{\circ}$] $\downarrow$& 21/0.47 & 15/0.33 & \textbf{11/0.27} \\
        \bottomrule
    \end{tabular}
    }
\end{table}
{
    \small
    \bibliographystyle{ieeenat_fullname}
    \bibliography{main}
}

\end{document}